\pdfoutput=1
\documentclass[conference]{IEEEtran}
\IEEEoverridecommandlockouts

\usepackage{cite}
\usepackage{amsmath,amssymb,amsfonts}
\usepackage{algorithm}
\usepackage{algorithmic}

\usepackage{graphicx}
\usepackage{textcomp}
\usepackage{url}
\usepackage{xcolor}
\usepackage{tikz}
\usepackage{nicematrix}

\usepackage{soul}

\newcommand{\vtheta}{\boldsymbol{\theta}}
\newcommand{\vJ}{\mathbf{J}}
\newcommand{\vv}[1]{\mathbf{#1}}
\newcommand{\norm}[1]{\left\|#1\right\|}
\newcommand{\abs}[1]{\left|#1\right|}
\DeclareMathOperator{\diag}{diag}
\DeclareMathOperator{\sign}{sign}

\begin{document}

\title{Scalable Levenberg-Marquardt Training Framework for Deep Learning}

\author{\IEEEauthorblockN{Xianghao Wang}
    \IEEEauthorblockA{\textit{School of Computing} \\
        \textit{Australian National University}\\
        Canberra, Australia \\
        xianghao.wang@anu.edu.au}
    \and
    \IEEEauthorblockN{Johan Barth\'elemy}
    \IEEEauthorblockA{\textit{NVIDIA} \\
        CA, United States \\
        jbarthelemy@nvidia.com
    }
    \and
    \IEEEauthorblockN{John Taylor}
    \IEEEauthorblockA{\textit{School of Computing} \\
        \textit{Australian National University}\\
        Canberra, Australia \\
        john.taylor@anu.edu.au
    }
}

\maketitle

\begin{abstract}
    The Levenberg-Marquardt (LMA) algorithm is a widely used nonlinear least-squares solver in scientific computing, known for its rapid convergence. Recent studies show that for small-scale neural networks, LMA provides better convergence and architectural robustness than first-order optimizers like Adam. However, as a second-order method, LMA imposes prohibitive memory and computational costs. These costs are primarily from the Jacobian matrix, whose dimensions scale with both the number of samples and parameters. This bottleneck limits LMA in its traditional applications and prevents its use in large-scale neural networks.

    Recent efforts scale LMA using preconditioned conjugate gradient (PCG) solvers to avoid materializing the Jacobian. However, PCG only approximates the solution, and an accurate approximation requires many network propagations and a task-specific preconditioner. Other studies optimize LMA to reduce costs, but they restrict their methods to specific tasks or network architectures. As a result, LMA remains impractical for training general large-scale neural networks.

    In this paper, we propose LeMA, a distributed LMA algorithm for large-scale neural network training on multi-GPU platforms. The contributions of LeMA are twofold. First, an exact dual formulation of LMA removes the memory dependence on the parameter dimension. Second, a Jacobian sharding strategy computes Jacobian shards on the fly and in parallel to avoid materializing the full Jacobian.

    For large-scale neural networks, LeMA reduces the memory footprint of LMA from quadratic in the number of parameters to quadratic in the batch size. This advancement makes exact LMA steps feasible for billion-parameter models without approximate PCG solves. Furthermore, for a convolutional neural network with 1,000,022,632 parameters on an 8-GPU node, LeMA achieves a $7.97\times$ speedup over a single GPU, corresponding to a 99.7\% strong scaling efficiency.
\end{abstract}
\begin{IEEEkeywords}
    Levenberg-Marquardt algorithm, distributed training, deep learning, multi-GPU
\end{IEEEkeywords}

\section{Introduction}\label{sec:introduction}
The Levenberg-Marquardt (LMA) algorithm~\cite{levenberg1944method,marquardt1963algorithm} is a standard solver for nonlinear least-squares problems in scientific computing~\cite{nocedal2006numerical,gavin2019levenberg}. Each LMA step solves a damped Gauss-Newton system. A damping factor $\lambda$ moves the step between the Gauss-Newton method and gradient descent, allowing LMA to converge quickly near a minimum while remaining robust far from it. Since Hagan and Menhaj~\cite{hagan1994training} applied LMA to neural networks, many studies have used it to train small models containing up to several thousand parameters~\cite{ali2023levenberg,zhang2023lifetime,taylor2022optimizing,revanesh2024artificial,troiano2024comparative,koshy2024lmhistnet,dimarco2025torch-levenberg,dimarco2025tf-levenberg}.

Despite its fast convergence, LMA is difficult to scale to large neural networks. The bottleneck lies in the linear system solved at each step, which involves an $M \times M$ coefficient matrix $\vJ^T\vJ+\lambda \operatorname{diag}(\vJ^T\vJ)$ and an $N \times M$ Jacobian matrix $\vJ$, where $N$ is the batch size and $M$ is the number of parameters. This formulation creates an $\mathcal{O}(MN+M^2)$ memory footprint. For a model with only one million parameters, the coefficient matrix alone contains a trillion elements. The computational cost is equally high. Evaluating $\vJ$ requires $N$ backward propagations, computing the coefficient matrix takes $\mathcal{O}(M^2 N)$ time, and solving the linear system takes $\mathcal{O}(M^3)$ time. These extreme memory and computation demands prevent the extension of LMA to large-scale training.

To address these memory bottlenecks, several studies and libraries~\cite{jaxopt_implicit_diff,kandel2021efficient,hollein20253dgs,petsc-web-page,symoens2025inexact} use preconditioned conjugate gradient (PCG) solvers~\cite{hestenes1952methods} to avoid materializing the large matrices. PCG computes an approximate solution to the linear system $\vv{A}\vv{x} = \vv{b}$ using only matrix-vector products $\vv{A}\vv{v}$, without explicitly constructing $\vv{A}$. For instance, both the optimization toolkit \texttt{PETSc}~\cite{petsc-web-page} and the Python library \texttt{JAXopt}~\cite{jaxopt_implicit_diff} provide LMA implementations with a PCG option. However, applying PCG to LMA requires numerous network propagations to ensure a robust approximation. Additionally, PCG relies on a suitable preconditioner matrix, and selecting an effective one is highly task-specific~\cite{agarwal2010bundle}.

As an exact alternative to PCG, Bilski et al.~\cite{bilski2020local} decompose LMA into independent tasks where each neuron computes its own Jacobian and weight updates. However, this approach assumes each neuron produces a single output. It is therefore incompatible with convolutional layers, which share parameters to generate multiple spatial outputs.

Our goal is to design an exact LMA framework for general deep learning tasks, without relying on PCG approximations or specific network architectures. We focus strictly on the systems engineering challenges to make each exact LMA step feasible and fast at scale, as the numerical convergence properties of LMA fall beyond the scope of this work. Di Marco~\cite{dimarco2025tf-levenberg,dimarco2025torch-levenberg} utilizes a dual form of the simpler Levenberg method, which applies $\lambda \vv{I}$ as the damping term. This approach successfully reduces the memory footprint from $\mathcal{O}(MN+M^2)$ to $\mathcal{O}(MN+N^2)$, making it feasible to train large neural networks where $M \gg N$. Inspired by this, we derive and implement an exact dual formulation specifically for LMA. To further reduce memory consumption, we design a partitioning strategy that avoids storing the full Jacobian $\vJ$. This design allows us to parallelize LMA and accelerate training across multiple GPUs.

We present \textit{LeMA}, a multi-GPU Levenberg-Marquardt training framework for neural networks. We summarize our main contributions as follows:
\begin{itemize}
    \item We derive an exact dual formulation for LMA, which shifts the linear system from the parameter dimension to the batch dimension (Section~\ref{sec:algorithm-dual-form}).
    \item We provide guidelines for designing robust residual functions tailored for LMA (Section~\ref{sec:algorithm-residual-derivation}).
    \item We design a partitioning and parallelization strategy to further reduce the memory footprint and accelerate computation (Sections~\ref{sec:implementation-partition} and~\ref{sec:implementation-jacobian-shard-computation}).
\end{itemize}
In addition to these core contributions, we introduce several supplementary optimizations detailed in Sections~\ref{sec:algorithm} and~\ref{sec:implementation}.

We evaluate LeMA on image classification tasks across various neural networks. LeMA scales LMA to models with up to $1{,}000{,}022{,}632$ parameters and achieves a $99.7\%$ strong scaling efficiency on 8 GPUs. For LeNet-5, which is small enough for the standard LMA form to fit in GPU memory, LeMA achieves an $8.3\times$ speedup and a $33\times$ reduction in memory footprint on a single GPU compared to a standard LMA implementation. When running on 8 GPUs, it delivers a $37.8\times$ speedup and a $41\times$ memory reduction over a standard multi-GPU LMA baseline.

The rest of this paper is organized as follows. Section~\ref{sec:background} reviews LMA and Jacobian computation within deep learning frameworks. Section~\ref{sec:algorithm} derives the exact dual formulation and provides residual representations for common loss functions. Section~\ref{sec:implementation} details the distributed implementation of LeMA and analyzes its associated costs. Finally, Section~\ref{sec:result} evaluates the performance of LeMA.
\section{Background}\label{sec:background}

This section reviews the Levenberg-Marquardt algorithm (Section~\ref{sec:background-levenberg}) and the computation of Jacobian matrices within deep learning frameworks (Section~\ref{sec:background-jacobian}).

\subsection{The Levenberg-Marquardt Algorithm}\label{sec:background-levenberg}

The Levenberg-Marquardt (LMA) algorithm solves nonlinear least-squares problems~\cite{levenberg1944method,marquardt1963algorithm,nocedal2006numerical,gavin2019levenberg}. Consider a model $f(\cdot; \vtheta)$ with parameters $\vtheta \in \mathbb{R}^M$ and a batch of $N$ samples with inputs $\vv{X} \in \mathbb{R}^{N \times d_\text{in}}$ and targets $\vv{Y} \in \mathbb{R}^{N \times d_\text{out}}$, where the $i$-th rows correspond to $\vv{x}_i$ and $\vv{y}_i$. LMA minimizes the sum of squared residuals $\norm{\vv{r}}_2^2$, where $\vv{r} \in \mathbb{R}^N$ contains one scalar residual per sample. In classical curve fitting, the model output is a scalar, making the residual simply $r_i = f(\vv{x}_i; \vtheta) - y_i$. However, neural networks typically produce vector outputs, such as a probability distribution over classes. Following Hagan and Menhaj~\cite{hagan1994training}, recent works~\cite{dimarco2025torch-levenberg,revanesh2024artificial,ali2023levenberg,zhang2023lifetime,troiano2024comparative,koshy2024lmhistnet} therefore define each residual as $r_i = r(f(\vv{x}_i; \vtheta); \vv{y}_i)$, where the residual function $r$ maps the $d_\text{out}$-dimensional output of a sample and its target to a scalar. Section~\ref{sec:algorithm-residual-derivation} derives $r$ for common loss functions.

Each LMA step computes the parameter update $\Delta \vtheta$ by solving
\begin{equation}\label{eq:levenberg-marquardt-residual-equation}
  (\vJ^T \vJ + \lambda \vv{D}) \Delta \vtheta = -\vJ^T \vv{r}, \quad \vv{D} = \diag(\vJ^T \vJ)
\end{equation}
where $\vJ = \partial \vv{r} / \partial \vtheta \in \mathbb{R}^{N \times M}$ is the Jacobian with its $i$-th row defined as $(\nabla_{\vtheta} r_i)^T$, and $\lambda > 0$ is the \textit{damping factor}. Decreasing $\lambda$ moves LMA towards the Gauss-Newton method for faster convergence, whereas increasing $\lambda$ moves it towards gradient descent for robustness.

Implementations of LMA differ in their update criteria, damping strategies, and iteration logic. We adopt the logic of \texttt{torch-levenberg-marquardt}~\cite{dimarco2025torch-levenberg}, a Python library for training neural networks using LMA, as detailed in Algorithm~\ref{alg:levenberg-marquardt}. A step is accepted only if it reduces the sum of squared residuals, specifically when $\norm{\vv{r}'}_2^2 < \norm{\vv{r}}_2^2$. In this case, $\lambda$ is decreased; otherwise, $\lambda$ is increased and the step is recomputed.

\begin{algorithm}[t]
  \caption{The Levenberg-Marquardt Algorithm}
  \label{alg:levenberg-marquardt}
  \begin{algorithmic}[1]
    \REQUIRE Model $f(\cdot; \vtheta)$, initial parameters $\vtheta_0$, data $(\vv{X}, \vv{Y})$, initial damping factor $\lambda_0$, adjustment factors $\lambda_{\uparrow} > 1$, $0 < \lambda_{\downarrow} < 1$, maximum iterations $T$
    \ENSURE Trained parameters $\vtheta$
    \STATE $\vtheta \leftarrow \vtheta_0$, $\lambda \leftarrow \lambda_0$
    \FOR{$t = 1$ \TO $T$}
    \STATE $\vv{r} \leftarrow \vv{r}(f(\vv{X}; \vtheta); \vv{Y})$ \COMMENT{residuals}
    \STATE $\vJ \leftarrow \partial \vv{r} / \partial \vtheta$ \COMMENT{Jacobian}
    \STATE $\vv{D} \leftarrow \diag(\vJ^T \vJ)$
    \REPEAT
    \STATE Solve $(\vJ^T \vJ + \lambda \vv{D}) \Delta \vtheta = -\vJ^T \vv{r}$ for $\Delta \vtheta$
    \STATE $\vtheta' \leftarrow \vtheta + \Delta \vtheta$
    \STATE $\vv{r}' \leftarrow \vv{r}(f(\vv{X}; \vtheta'); \vv{Y})$
    \IF{$\|\vv{r}'\|^2_2 < \|\vv{r}\|^2_2$}
    \STATE $\vtheta \leftarrow \vtheta'$, $\lambda \leftarrow \lambda \cdot \lambda_{\downarrow}$ \COMMENT{accept step}
    \ELSE
    \STATE $\lambda \leftarrow \lambda \cdot \lambda_{\uparrow}$ \COMMENT{reject step}
    \ENDIF
    \UNTIL{step accepted}
    \ENDFOR
    \RETURN $\vtheta$
  \end{algorithmic}
\end{algorithm}

Many studies~\cite{troiano2024comparative,bilski2020local,hagan1994training} and applications~\cite{dimarco2025torch-levenberg,ali2023levenberg,revanesh2024artificial,zhang2023lifetime} confuse LMA with the earlier Levenberg algorithm~\cite{levenberg1944method}, which damps all parameters equally with the identity matrix $\vv{I}_M$:
\begin{equation}\label{eq:levenberg-equation}
  (\vJ^T \vJ + \lambda \vv{I}_M) \Delta \vtheta = -\vJ^T \vv{r}
\end{equation}
Marquardt~\cite{marquardt1963algorithm} replaced $\vv{I}_M$ with $\vv{D} = \diag(\vJ^T \vJ)$, which scales the damping of each parameter according to its curvature.

\subsection{Jacobian Matrix Computation}\label{sec:background-jacobian}

Deep learning frameworks~\cite{Ansel_PyTorch_2_Faster_2024,jax2018github,Abadi_TensorFlow_Large-scale_machine_2015} provide two primitives to compute products involving the Jacobian without explicitly materializing the full matrix: the \textit{vector-Jacobian product (VJP)} and the \textit{Jacobian-vector product (JVP)}. A VJP computes $\vJ^T \vv{v}$ for a vector $\vv{v} \in \mathbb{R}^N$; setting $\vv{v}$ to the standard basis vector $\vv{e}_i$ extracts the $i$-th row of $\vJ$. A JVP computes $\vJ\vv{v}$ for a vector $\vv{v} \in \mathbb{R}^M$; setting $\vv{v} = \vv{e}_j$ extracts the $j$-th column. Equation~\eqref{eq:jacobian-residual-full} illustrates both operations. Internally, a VJP relies on reverse-mode automatic differentiation, requiring a forward pass that stores intermediate activations followed by a backward pass. A JVP instead relies on forward-mode automatic differentiation, requiring only a single forward pass without storing activations.

\begin{equation}\label{eq:jacobian-residual-full}
  \vJ = \begin{bNiceMatrix}[last-row, last-col]
    \frac{\partial r_1}{\partial \theta_1} & \cdots & \frac{\partial r_1}{\partial \theta_j}          & \cdots & \frac{\partial r_1}{\partial \theta_M} &                                         \\
    \vdots                                 & \ddots & \vdots                                          &        & \vdots                                 &                                         \\
    \frac{\partial r_i}{\partial \theta_1} & \cdots & \frac{\partial r_i}{\partial \theta_j}          & \cdots & \frac{\partial r_i}{\partial \theta_M} & \textcolor{blue}{\leftarrow \text{VJP}} \\
    \vdots                                 &        & \vdots                                          & \ddots & \vdots                                 &                                         \\
    \frac{\partial r_N}{\partial \theta_1} & \cdots & \frac{\partial r_N}{\partial \theta_j}          & \cdots & \frac{\partial r_N}{\partial \theta_M} &                                         \\
                                           &        & \textcolor{red}{\overset{\uparrow}{\text{JVP}}} &        &                                        &
    \CodeAfter
    \tikz \draw [blue, rounded corners] (3-|1) rectangle (4-|6);
    \tikz \draw [red, rounded corners] (1-|3) rectangle (6-|4);
  \end{bNiceMatrix} \in \mathbb{R}^{N \times M}
\end{equation}

Our framework assumes that each sample in a batch is processed independently, which excludes intra-batch operations such as Batch Normalization. Under this assumption, $r_i$ depends only on the $i$-th sample. Therefore, one single-sample VJP yields one row of $\vJ$, and assembling the full Jacobian requires $N$ single-sample VJPs. In contrast, a column involves all $N$ samples, meaning its extraction requires $N$ single-sample JVPs, or equivalently, one $N$-sample JVP. Assembling the full Jacobian column-wise thus requires $MN$ single-sample JVPs. Consequently, we always construct the Jacobian row-wise using VJPs.
\section{Algorithm Design}\label{sec:algorithm}

This section introduces the core algorithm of LeMA. Specifically, Section~\ref{sec:algorithm-dual-form} derives an exact dual formulation of LMA to bypass the $M \times M$ matrix $\vJ^T\vJ$. Section~\ref{sec:algorithm-residual-derivation} formulates residuals for common loss functions, and Section~\ref{sec:algorithm-lema} outlines the complete LeMA algorithm.

\subsection{Dual Transformation}\label{sec:algorithm-dual-form}

When training large-scale models where $M > N$, the dense $M \times M$ Gram matrix $\vJ^T\vJ$ in~\eqref{eq:levenberg-marquardt-residual-equation} dominates the memory footprint. We therefore transform the LMA equation into its dual form, which only involves the $N \times N$ matrix $\vJ\vv{D}^{-1}\vJ^T$ and thus moves the linear system from the parameter dimension to the batch dimension:
\begin{equation}\label{eq:dual-levenberg-marquardt-equation}
    (\vJ\vv{D}^{-1} \vJ^T + \lambda \vv{I}_N) \vv{v} = -\vv{r}, \quad \Delta \vtheta = \vv{D}^{-1}\vJ^T \vv{v}
\end{equation}
where $\vv{D} = \diag(\vJ^T\vJ)$ as in~\eqref{eq:levenberg-marquardt-residual-equation}. We assume that $\vv{D}$ is invertible. Section~\ref{sec:implementation-standard} describes how we ensure this in practice.

We now prove that~\eqref{eq:dual-levenberg-marquardt-equation} yields the same update as~\eqref{eq:levenberg-marquardt-residual-equation}. Since $\lambda > 0$ and $\vv{D}$ has a positive diagonal, both $\vJ^T\vJ + \lambda \vv{D}$ and $\vJ\vv{D}^{-1}\vJ^T + \lambda \vv{I}_N$ are positive definite and thus invertible. Expanding the products shows that
\begin{equation*}
    \begin{aligned}
        (\vJ^T\vJ + \lambda \vv{D})\vv{D}^{-1} \vJ^T & = \vJ^T\vJ\vv{D}^{-1} \vJ^T + \lambda \vJ^T      \\
                                                     & = \vJ^T(\vJ\vv{D}^{-1} \vJ^T + \lambda \vv{I}_N)
    \end{aligned}
\end{equation*}
Multiplying both sides by $(\vJ^T\vJ + \lambda \vv{D})^{-1}$ on the left and by $-(\vJ\vv{D}^{-1} \vJ^T + \lambda \vv{I}_N)^{-1}\vv{r}$ on the right turns the update of~\eqref{eq:levenberg-marquardt-residual-equation} into that of~\eqref{eq:dual-levenberg-marquardt-equation}:
\begin{equation*}
    \begin{aligned}
        \Delta \vtheta & = -(\vJ^T\vJ + \lambda \vv{D})^{-1} \vJ^T \vv{r}                                                     \\
                       & = -\vv{D}^{-1} \vJ^T (\vJ\vv{D}^{-1} \vJ^T + \lambda \vv{I}_N)^{-1} \vv{r} = \vv{D}^{-1}\vJ^T \vv{v}
    \end{aligned}
\end{equation*}

Notably, the dual formulation in recent implementations~\cite{dimarco2025torch-levenberg,dimarco2025tf-levenberg} corresponds to the Levenberg algorithm~\eqref{eq:levenberg-equation}, i.e., $\vv{D} = \vv{I}_M$ in~\eqref{eq:dual-levenberg-marquardt-equation}, and thus lacks Marquardt's defining modification $\vv{D} = \diag(\vJ^T\vJ)$~\cite{marquardt1963algorithm}. In contrast, we derive and prove an exact dual formulation of the complete LMA algorithm.

\subsection{Residual Formulation}\label{sec:algorithm-residual-derivation}

Deep learning models minimize a standard loss $\mathcal{L} = \frac{1}{N}\sum_{i=1}^{N} \ell_i$, whereas LMA minimizes the sum of squared residuals $\norm{\vv{r}}_2^2 = \sum_{i=1}^{N} r_i^2$~\cite{hagan1994training}. As a practical heuristic to guide the migration of existing code to the LMA framework, Equation~\eqref{eq:residuals} formulates the exact residual representations for three common loss functions. Here, $\hat{y}_i$ denotes the network output for the $i$-th sample, while $\hat{y}_{i,c}$ and $y_{i,c}$ correspond to the $c$-th of $C$ classes:
\begin{equation}\label{eq:residuals}
    \begin{array}{l@{\quad}l@{\quad}l}
        \text{Loss} & \ell_i                                    & r_i                                            \\
        \text{MSE}  & (\hat{y}_i - y_i)^2                       & \hat{y}_i - y_i                                \\
        \text{CE}   & -\sum_{c=1}^{C} y_{i,c}\log \hat{y}_{i,c} & \sqrt{\ell_i + \epsilon}                       \\
        \text{MAE}  & \abs{\hat{y}_i - y_i}                     & \sign(\hat{y}_i - y_i)\sqrt{\ell_i + \epsilon}
    \end{array}
\end{equation}

The Mean Squared Error (MSE) directly yields a residual. For the Cross-Entropy (CE) loss, the natural residual would be $\sqrt{\ell_i}$. However, the derivative of $\sqrt{x}$ is $\frac{1}{2\sqrt{x}}$, so a perfectly predicted sample ($\ell_i = 0$) causes a division by zero in the Jacobian computation. A small constant $\epsilon > 0$ inside the square root prevents this singularity.

The Mean Absolute Error (MAE) requires $\epsilon > 0$ for the same reason, and additionally the sign of the error $e_i = \hat{y}_i - y_i$. Without the sign, the derivative of $\sqrt{\abs{e_i} + \epsilon}$ with respect to $e_i$ is $\frac{\sign(e_i)}{2\sqrt{\abs{e_i} + \epsilon}}$, which jumps between $-\frac{1}{2\sqrt{\epsilon}}$ and $\frac{1}{2\sqrt{\epsilon}}$ as $e_i$ crosses zero. These values are large for a small $\epsilon$, so the Jacobian becomes discontinuous, and the optimizer oscillates around the minimum. Multiplying the residual by $\sign(e_i)$ cancels the sign in the derivative, which becomes $\frac{1}{2\sqrt{\abs{e_i} + \epsilon}}$, i.e., strictly positive and continuous at $e_i = 0$.

In summary, matching a loss to the LMA objective often introduces a square root into the residual and hence into the denominator of its derivative. Residual functions must therefore avoid near-zero denominators and sign oscillations as the error approaches zero.

\subsection{The LeMA Algorithm}\label{sec:algorithm-lema}

Algorithm~\ref{alg:levenberg-marquardt-lema} presents the complete LeMA procedure. We collectively refer to $\vJ^T\vJ$ in the standard form~\eqref{eq:levenberg-marquardt-residual-equation} and $\vJ\vv{D}^{-1}\vJ^T$ in the dual form~\eqref{eq:dual-levenberg-marquardt-equation} as the Gram matrix $\vv{G}$. LeMA selects the dual form when $M > N$ and the standard form otherwise, so $\vv{G}$ has $\min(M, N)^2$ elements. The memory footprint thus drops from $\mathcal{O}(MN + M^2)$ in Algorithm~\ref{alg:levenberg-marquardt} to $\mathcal{O}(MN + \min(M, N)^2)$, and Section~\ref{sec:implementation-partition} further eliminates the $\mathcal{O}(MN)$ term of the Jacobian. In addition, LeMA forms $\vv{G}$ once per iteration and reuses it across rejected steps, which only change $\lambda$, and carries the residuals of an accepted step over to the next iteration.

\begin{algorithm}[t]
    \caption{The Levenberg-Marquardt Algorithm of LeMA}
    \label{alg:levenberg-marquardt-lema}
    \begin{algorithmic}[1]
        \REQUIRE Model $f(\cdot; \vtheta)$ with $M$ parameters, per-sample residual function $r$, initial parameters $\vtheta_0$, data $(\vv{X}, \vv{Y})$ with $N$ samples, initial damping factor $\lambda_0$, adjustment factors $\lambda_{\uparrow} > 1$, $0 < \lambda_{\downarrow} < 1$, maximum iterations $T$
        \ENSURE Trained parameters $\vtheta$, loss value $\mathcal{L}$
        \STATE $\vtheta \leftarrow \vtheta_0$, $\lambda \leftarrow \lambda_0$
        \STATE $\vv{r} \leftarrow \vv{r}(f(\vv{X}; \vtheta); \vv{Y})$ \COMMENT{residuals, Sec.~\ref{sec:algorithm-residual-derivation}}
        \FOR{$t = 1$ \TO $T$}
        \STATE $\vJ \leftarrow \partial \vv{r} / \partial \vtheta$ \COMMENT{Jacobian}
        \STATE $\vv{D} \leftarrow \diag(\vJ^T \vJ)$
        \IF{$M > N$}
        \STATE $\vv{G} \leftarrow \vJ \vv{D}^{-1} \vJ^T$ \COMMENT{dual form, $N \times N$}
        \ELSE
        \STATE $\vv{G} \leftarrow \vJ^T \vJ$, $\vv{g} \leftarrow \vJ^T \vv{r}$ \COMMENT{standard form, $M \times M$}
        \ENDIF
        \REPEAT
        \IF{$M > N$}
        \STATE Solve $(\vv{G} + \lambda \vv{I}_N) \vv{v} = -\vv{r}$ for $\vv{v}$
        \STATE $\Delta \vtheta \leftarrow \vv{D}^{-1} \vJ^T \vv{v}$
        \ELSE
        \STATE Solve $(\vv{G} + \lambda \vv{D}) \Delta \vtheta = -\vv{g}$ for $\Delta \vtheta$
        \ENDIF
        \STATE $\vtheta' \leftarrow \vtheta + \Delta \vtheta$
        \STATE $\vv{r}' \leftarrow \vv{r}(f(\vv{X}; \vtheta'); \vv{Y})$
        \IF{$\|\vv{r}'\|^2 < \|\vv{r}\|^2$}
        \STATE $\vtheta \leftarrow \vtheta'$, $\vv{r} \leftarrow \vv{r}'$, $\lambda \leftarrow \lambda \cdot \lambda_{\downarrow}$ \COMMENT{accept step}
        \ELSE
        \STATE $\lambda \leftarrow \lambda \cdot \lambda_{\uparrow}$ \COMMENT{reject step}
        \ENDIF
        \UNTIL{step accepted}
        \ENDFOR
        \STATE $\mathcal{L} \leftarrow \frac{1}{N} \norm{\vv{r}}_2^2$
        \RETURN $\vtheta$, $\mathcal{L}$
    \end{algorithmic}
\end{algorithm}

\section{LeMA Implementation}\label{sec:implementation}

This section describes how LeMA executes Algorithm~\ref{alg:levenberg-marquardt-lema} on $P$ GPUs. Section~\ref{sec:implementation-setup} covers the distributed setup, Section~\ref{sec:implementation-partition} the partitioned computation of both forms, Section~\ref{sec:implementation-jacobian-shard-computation} the evaluation of Jacobian shards, and Section~\ref{sec:implementation-analysis} the resulting cost.

\subsection{Distributed Setup}\label{sec:implementation-setup}

LeMA runs $P$ processes in the Single Program, Multiple Data (SPMD) model~\cite{Darema2011}, one per NVIDIA GPU, and communicates through \texttt{torch.distributed}~\cite{Ansel_PyTorch_2_Faster_2024}. Each process belongs to two communication groups. Large tensors move between GPUs over NCCL~\cite{hu2025demystifying}, NVIDIA's collective communication library, and small scalars such as local loss values move between hosts over Meta's Gloo library~\cite{meta2025gloo}. Keeping the scalars on the host avoids many small host-to-device and device-to-host copies that would otherwise interrupt the GPU streams.

All parameters are stored in one flat tensor, and the weights of each layer are mapped to their positions in it. A VJP therefore returns a single $M$-dimensional gradient, and no per-layer gradients need to be concatenated.

Each process loads its own mini-batch of $N/P$ samples instead of receiving part of a scattered global batch, and the $P$ mini-batches together form the global batch. PyTorch's \texttt{DistributedSampler}~\cite{Ansel_PyTorch_2_Faster_2024} keeps the mini-batches disjoint. The dual form also needs the global batch on every process (Section~\ref{sec:implementation-dual}).


\subsection{Partition and Parallelization}\label{sec:implementation-partition}

The $N \times M$ Jacobian dominates the memory of Algorithm~\ref{alg:levenberg-marquardt-lema}, so we never store it in full. We split $\vJ$ by rows into $N/S$ \textit{shards} $\vJ = [\vJ_1^T, \dots, \vJ_{N/S}^T]^T$, where $\vJ_k \in \mathbb{R}^{S \times M}$ and $\vv{r}_k \in \mathbb{R}^S$ correspond to the $k$-th group of $S$ samples. Each shard is computed on the fly, added to the Gram matrix $\vv{G}$, and freed, so the Jacobian occupies $\mathcal{O}(SM)$ memory instead of $\mathcal{O}(NM)$. We split by rows because a row shard needs only $S$ single-sample VJPs. The subroutine $\textsc{Jac}$ evaluates a shard and returns $\vJ_k$ and $\vv{r}_k$, and we drop $\vv{r}_k$ from the notation when it is not used. Internally, $\textsc{Jac}$ works on \textit{slices} of $L$ samples (Section~\ref{sec:implementation-jacobian-shard-computation}). On process $p$, $(\vv{X}_{p,i}, \vv{Y}_{p,i})$ and $\vJ_{p,i}$ denote the $i$-th of the $N/(PS)$ shards of its mini-batch $(\vv{X}_p, \vv{Y}_p)$. We assume that $PS$ divides $N$ and $L$ divides $S$.

\subsubsection{Standard Form}\label{sec:implementation-standard}
Under the row partition, both terms of~\eqref{eq:levenberg-marquardt-residual-equation} become sums over shards:
\begin{equation}\label{eq:compute-standard-shard}
    \vJ^T\vJ = \sum_{k=1}^{N/S} \vJ_k^T\vJ_k, \qquad \vJ^T\vv{r} = \sum_{k=1}^{N/S} \vJ_k^T\vv{r}_k
\end{equation}
The summands are independent, so each process sums the shards of its own mini-batch locally. An NCCL all-reduce then gives every process $\vJ^T\vJ$ and $\vJ^T\vv{r}$, and a Gloo all-reduce gives $\norm{\vv{r}}_2^2$. Parameters without curvature, such as those of inactive ReLU units, have a zero diagonal entry and would make $\vv{D}$ singular. We therefore clamp each $(\vJ^T\vJ)_{jj}$ in $\vv{D}$ to at least $1\%$ of the mean diagonal entry, and the dual form uses the same clamp. Every process then solves the $M \times M$ system on its own. Algorithm~\ref{alg:standard-distributed} lists the steps on process $p$. If LeMA rejects a step, only the final solve is redone with the new $\lambda$.

\begin{algorithm}[t]
    \caption{Distributed Standard Form Solve on Process $p$}
    \label{alg:standard-distributed}
    \begin{algorithmic}[1]
        \REQUIRE Local mini-batch $(\vv{X}_p, \vv{Y}_p)$ with shards $(\vv{X}_{p,i}, \vv{Y}_{p,i})$ of $S$ samples, parameters $\vtheta$, damping factor $\lambda$
        \ENSURE Parameter update $\Delta\vtheta$, loss $\norm{\vv{r}}_2^2$
        \STATE $\vv{G}_p \leftarrow \vv{0}^{M \times M}$, $\vv{g}_p \leftarrow \vv{0}^{M}$, $\rho_p \leftarrow 0$
        \FOR{$i = 1$ \TO $N/(PS)$}
        \STATE $\vJ_{p,i}, \vv{r}_{p,i} \leftarrow \textsc{Jac}(\vv{X}_{p,i}, \vv{Y}_{p,i})$ \COMMENT{$\vJ_{p,i} \in \mathbb{R}^{S \times M}$}
        \STATE $\vv{G}_p \leftarrow \vv{G}_p + \vJ_{p,i}^T \vJ_{p,i}$
        \STATE $\vv{g}_p \leftarrow \vv{g}_p + \vJ_{p,i}^T \vv{r}_{p,i}$
        \STATE $\rho_p \leftarrow \rho_p + \vv{r}_{p,i}^T \vv{r}_{p,i}$
        \ENDFOR
        \STATE $\vJ^T\vJ \leftarrow \textsc{AllReduce}(\vv{G}_p)$, $\vJ^T\vv{r} \leftarrow \textsc{AllReduce}(\vv{g}_p)$ \COMMENT{NCCL}
        \STATE $\norm{\vv{r}}_2^2 \leftarrow \textsc{AllReduce}(\rho_p)$ \COMMENT{Gloo}
        \STATE $\vv{D} \leftarrow \diag(\vJ^T\vJ)$
        \STATE Solve $(\vJ^T\vJ + \lambda \vv{D}) \Delta\vtheta = -\vJ^T\vv{r}$ for $\Delta\vtheta$
        \RETURN $\Delta\vtheta$, $\norm{\vv{r}}_2^2$
    \end{algorithmic}
\end{algorithm}

\subsubsection{Dual Form}\label{sec:implementation-dual}
The dual form~\eqref{eq:dual-levenberg-marquardt-equation} first solves $(\vJ\vv{D}^{-1}\vJ^T + \lambda \vv{I}_N)\vv{v} = -\vv{r}$ and then sets $\Delta\vtheta = \vv{D}^{-1}\vJ^T\vv{v}$. To avoid forming any $M \times M$ matrix, we keep only the diagonal of $\vv{D}^{-1}$ as a vector $\vv{d} \in \mathbb{R}^M$ with $d_j = 1/(\vJ^T\vJ)_{jj}$. Multiplying by $\vv{D}^{-1}$ on the right then scales column $j$ by $d_j$, so $\tilde{\vJ}_k = \vJ_k\vv{D}^{-1} = \vJ_k \odot \vv{d}^T$, where $\odot$ is element-wise multiplication and $\vv{d}^T$ is broadcast across rows. Under the row partition, the Gram matrix splits into $S \times S$ blocks:
\begin{equation}\label{eq:compute-dual-shard}
    \vJ\vv{D}^{-1}\vJ^T =
    \begin{bmatrix}
        \tilde{\vJ}_1\vJ_1^T     & \cdots & \tilde{\vJ}_1\vJ_{N/S}^T     \\
        \vdots                   & \ddots & \vdots                       \\
        \tilde{\vJ}_{N/S}\vJ_1^T & \cdots & \tilde{\vJ}_{N/S}\vJ_{N/S}^T
    \end{bmatrix}
\end{equation}

Algorithm~\ref{alg:dual-distributed} computes the dual form on process $p$ in three phases. The first phase computes $\vv{d}$. Because $(\vJ^T\vJ)_{jj}$ is the squared norm of column $j$ of $\vJ$, the vector $\vv{d}$ can be accumulated shard by shard and all-reduced in the same way as $\vJ^T\vJ$ in Algorithm~\ref{alg:standard-distributed}. The diagonal entries are clamped as in the standard form before taking reciprocals. In the second phase, process $p$ computes the $N/P$ rows of~\eqref{eq:compute-dual-shard} that belong to its mini-batch. For each local shard $\tilde{\vJ}_{p,i}$, it evaluates every global shard $\vJ_j$ and the block $\tilde{\vJ}_{p,i}\vJ_j^T$, and the global residuals $\vv{r}_j$ come out of the same evaluation. An all-gather then assembles the full Gram matrix on every process. In the third phase, every process solves for $\vv{v}$, and $\textsc{Vjp}$ computes $\vJ^T\vv{v}$ slice by slice over the local mini-batch without forming the Jacobian again. Here $\vv{v}_{p,l}$ is the part of $\vv{v}$ for the $l$-th local slice. An all-reduce of the partial sums, followed by scaling with $\vv{d}$, gives $\Delta\vtheta$. If LeMA rejects a step, only the solve and the third phase are repeated with the new $\lambda$. The standard form evaluates each Jacobian row once, whereas the dual form re-evaluates every global shard for each local shard and thus spends extra computation to save memory (Section~\ref{sec:implementation-analysis}).

\begin{algorithm}[t]
    \caption{Distributed Dual Form Solve on Process $p$}
    \label{alg:dual-distributed}
    \begin{algorithmic}[1]
        \REQUIRE Global batch $(\vv{X}, \vv{Y})$ with shards $(\vv{X}_j, \vv{Y}_j)$ of $S$ samples, local mini-batch $(\vv{X}_p, \vv{Y}_p)$ with shards $(\vv{X}_{p,i}, \vv{Y}_{p,i})$ of $S$ samples and slices $(\vv{X}_{p,l}, \vv{Y}_{p,l})$ of $L$ samples, parameters $\vtheta$, damping factor $\lambda$
        \ENSURE Parameter update $\Delta\vtheta$, loss $\norm{\vv{r}}_2^2$
        \STATE $\vv{d}_p \leftarrow \vv{0}^{M}$
        \FOR{$i = 1$ \TO $N/(PS)$}
        \STATE $\vJ_{p,i} \leftarrow \textsc{Jac}(\vv{X}_{p,i}, \vv{Y}_{p,i})$
        \STATE $\vv{d}_p \leftarrow \vv{d}_p + \mathbf{1}_S^T (\vJ_{p,i} \odot \vJ_{p,i})$
        \ENDFOR
        \STATE $\vv{d} \leftarrow \mathbf{1} \oslash \textsc{AllReduce}(\vv{d}_p)$ \COMMENT{NCCL}
        \STATE $\vv{G}_p \leftarrow \vv{0}^{(N/P) \times N}$ \COMMENT{local rows of $\vJ\vv{D}^{-1}\vJ^T$}
        \FOR{$i = 1$ \TO $N/(PS)$}
        \STATE $\tilde{\vJ}_{p,i} \leftarrow \textsc{Jac}(\vv{X}_{p,i}, \vv{Y}_{p,i}) \odot \vv{d}^T$
        \FOR{$j = 1$ \TO $N/S$}
        \STATE $\vJ_j, \vv{r}_j \leftarrow \textsc{Jac}(\vv{X}_j, \vv{Y}_j)$
        \STATE $\vv{G}_p[i, j] \leftarrow \tilde{\vJ}_{p,i} \vJ_j^T$
        \ENDFOR
        \ENDFOR
        \STATE $\vJ\vv{D}^{-1}\vJ^T \leftarrow \textsc{AllGather}(\vv{G}_p)$ \COMMENT{NCCL}
        \STATE $\vv{r} \leftarrow [\vv{r}_1; \dots; \vv{r}_{N/S}]$
        \STATE Solve $(\vJ\vv{D}^{-1}\vJ^T + \lambda \vv{I}_N) \vv{v} = -\vv{r}$ for $\vv{v}$
        \STATE $\vv{u}_p \leftarrow \vv{0}^{M}$
        \FOR{$l = 1$ \TO $N/(PL)$}
        \STATE $\vv{u}_p \leftarrow \vv{u}_p + \textsc{Vjp}(\vv{X}_{p,l}, \vv{Y}_{p,l}, \vv{v}_{p,l})$
        \ENDFOR
        \STATE $\Delta\vtheta \leftarrow \vv{d} \odot \textsc{AllReduce}(\vv{u}_p)$ \COMMENT{NCCL}
        \RETURN $\Delta\vtheta$, $\norm{\vv{r}}_2^2$
    \end{algorithmic}
\end{algorithm}

Each block product $\tilde{\vJ}_{p,i}\vJ_j^T$ multiplies two $S \times M$ shards and produces an $S \times S$ block. A general matrix multiplication (GEMM) usually maps one output tile to one thread block. When $S \ll M$, the output has few tiles, so only a few thread blocks run, each looping over the full inner dimension $M$, and most streaming multiprocessors (SMs) of the GPU sit idle. We address this with split-K~\cite{Thakkar_CUTLASS_2023}. LeMA cuts the inner dimension into chunks, multiplies matching chunks of the two shards into partial $S \times S$ blocks with one batched GEMM, and sums the partial blocks. The number of chunks is a multiple of the SM count, so the batched GEMM keeps all SMs busy. The chunks are strided views of the shards, so no data is copied, and the partial blocks are small. Shards with more rows already give the GEMM enough parallelism, so LeMA uses split-K only for shards with few rows.

\subsection{Jacobian Shard Computation}\label{sec:implementation-jacobian-shard-computation}

This section describes the functions $\textsc{Jac}$ and $\textsc{Vjp}$ used in Algorithms~\ref{alg:standard-distributed} and~\ref{alg:dual-distributed}. Samples are independent, so the $S$ rows of a shard come from $S$ single-sample VJPs. Running them one after another sends a single sample through the network at a time and leaves most of the GPU idle. $\textsc{Jac}$ instead vectorizes the VJPs of a slice of $L$ samples with \texttt{vmap}~\cite{Ansel_PyTorch_2_Faster_2024} and obtains an $L \times M$ Jacobian slice per call (Algorithm~\ref{alg:function-jac}). The per-sample function also returns the residual as an auxiliary output, so the residuals need no extra forward pass. Vectorizing $L$ VJPs, however, also stores $L$ times as many activations, so the slice size $L$ trades execution speed against memory.

We slice the computation of $\vJ^T\vv{v}$ in the dual form in the same way. One VJP over all $N/P$ local samples would compute it, but its forward pass would keep the activations of all of them at once. $\textsc{Vjp}$ therefore handles one slice of $L$ samples per call, and Algorithm~\ref{alg:dual-distributed} sums the results. Unlike $\textsc{Jac}$, $\textsc{Vjp}$ does not build a Jacobian slice. A single backward pass on the scalar $\vv{v}_l^T\vv{r}_l$ gives the vector $\vJ_l^T\vv{v}_l \in \mathbb{R}^M$ directly (Algorithm~\ref{alg:function-vjp}).

\begin{algorithm}[t]
    \caption{Jacobian Shard Evaluation $\textsc{Jac}(\vv{X}_k, \vv{Y}_k)$}
    \label{alg:function-jac}
    \begin{algorithmic}[1]
        \REQUIRE Shard $(\vv{X}_k, \vv{Y}_k)$ of $S$ samples with slices $(\vv{X}_{k,l}, \vv{Y}_{k,l})$ of $L$ samples, model $f(\cdot; \vtheta)$, residual function $r$
        \ENSURE Jacobian shard $\vJ_k \in \mathbb{R}^{S \times M}$, residuals $\vv{r}_k \in \mathbb{R}^{S}$
        \STATE $g(\vv{x}, \vv{y}) \leftarrow \big(\nabla_{\vtheta} r(f(\vv{x}; \vtheta); \vv{y}),\ r(f(\vv{x}; \vtheta); \vv{y})\big)$
        \FOR{$l = 1$ \TO $S / L$}
        \STATE $\vJ_{k,l}, \vv{r}_{k,l} \leftarrow \texttt{vmap}(g)(\vv{X}_{k,l}, \vv{Y}_{k,l})$ \COMMENT{$\vJ_{k,l} \in \mathbb{R}^{L \times M}$}
        \ENDFOR
        \RETURN $[\vJ_{k,1}; \dots; \vJ_{k,S/L}]$, $[\vv{r}_{k,1}; \dots; \vv{r}_{k,S/L}]$
    \end{algorithmic}
\end{algorithm}

\begin{algorithm}[t]
    \caption{Vector-Jacobian Product $\textsc{Vjp}(\vv{X}_l, \vv{Y}_l, \vv{v}_l)$}
    \label{alg:function-vjp}
    \begin{algorithmic}[1]
        \REQUIRE Slice $(\vv{X}_l, \vv{Y}_l)$ of $L$ samples, vector $\vv{v}_l \in \mathbb{R}^{L}$, model $f(\cdot; \vtheta)$, residual function $r$
        \ENSURE $\vJ_l^T \vv{v}_l \in \mathbb{R}^{M}$
        \STATE $\vv{r}_l \leftarrow r(f(\vv{X}_l; \vtheta); \vv{Y}_l)$ \COMMENT{forward pass}
        \RETURN $\nabla_{\vtheta} (\vv{v}_l^T \vv{r}_l)$ \COMMENT{backward pass}
    \end{algorithmic}
\end{algorithm}

\subsection{Memory Footprint and Performance Analysis}\label{sec:implementation-analysis}

We now analyze the per-device cost of the standard form (Algorithm~\ref{alg:standard-distributed}) and the dual form (Algorithm~\ref{alg:dual-distributed}). Let $A$ be the number of activation elements stored for one sample in a forward pass, and $F$ the FLOPs of one single-sample VJP, which consists of one forward and one backward pass. Because samples are independent, a vectorized pass over $L$ samples in $\textsc{Jac}$ or $\textsc{Vjp}$ stores $LA$ activation elements and costs $LF$ FLOPs. We leave out the input mini-batch, which every training method needs. Table~\ref{tab:cost-analysis} lists the remaining components. In the asymptotic bounds below, we drop a term only if a retained term strictly dominates it.

\begin{table}[t]
    \caption{Per-Device Cost of Algorithms~\ref{alg:standard-distributed} and~\ref{alg:dual-distributed}}
    \label{tab:cost-analysis}
    \centering
    \renewcommand{\arraystretch}{1.2}
    \begin{tabular}{lcc}
        \hline
        Component                   & Standard Form      & Dual Form            \\
        \hline
        \multicolumn{3}{l}{\textit{Memory (elements)}}                          \\
        Gram matrix                 & $3M^2$             & $3N^2$               \\
        Jacobian shards             & $2SM$              & $3SM$                \\
        Slice activations           & $LA$               & $LA$                 \\
        Vectors                     & $\mathcal{O}(M)$   & $\mathcal{O}(M + N)$ \\
        \hline
        \multicolumn{3}{l}{\textit{Computation (FLOPs)}}                        \\
        Forward and backward passes & $NF/P$             & $(3 + N/S)\,NF/P$    \\
        Shard products              & $2NM^2/P$          & $2N^2M/P$            \\
        Linear solve                & $\mathcal{O}(M^3)$ & $\mathcal{O}(N^3)$   \\
        \hline
        \multicolumn{3}{l}{\textit{Communication (elements)}}                   \\
        NCCL collectives            & $M^2 + M$          & $N^2 + 2M$           \\
        \hline
    \end{tabular}
\end{table}

\subsubsection{Memory Footprint}
The memory of both forms has three main parts. The first is the Gram matrix, which is $\vJ^T\vJ$ with $M^2$ elements in the standard form and the gathered $\vJ\vv{D}^{-1}\vJ^T$ with $N^2$ elements in the dual form. LeMA keeps the Gram matrix for reuse after a rejected step, so the solver applies damping to a separate copy and factorizes it into LU, which triples this cost. The second part is the Jacobian shards. $\textsc{Jac}$ computes the per-sample gradients slice by slice and then assembles them into the rows of the shard, so the gradients and their assembled copy briefly coexist and take $2SM$ elements. The standard form evaluates one shard at a time. The nested loop of the dual form also holds $\tilde{\vJ}_{p,i}$ while evaluating $\vJ_j$, which raises this part to $3SM$ elements. The third part is the activations of one slice, $LA$ elements, since $\textsc{Jac}$ and $\textsc{Vjp}$ never run at the same time. The remaining vectors, such as $\vtheta$, $\vv{d}$, and $\vv{r}$, are bounded by $SM$ and $N^2$. Algorithm~\ref{alg:levenberg-marquardt-lema} selects the dual form only when $M > N$, so the memory per device is
\begin{equation}\label{eq:memory-complexity}
    \mathcal{O}\big(\min(M, N)^2 + SM + LA\big)
\end{equation}
The first term is quadratic in the smaller of $M$ and $N$. The other two grow linearly with the model size and are set by the shard size $S$ and the slice size $L$, regardless of the batch size $N$. Algorithm~\ref{alg:levenberg-marquardt}, by contrast, stores the full $N \times M$ Jacobian.

\subsubsection{Computation}
In the standard form, each Jacobian row is evaluated exactly once, so each device runs $N/P$ single-sample VJPs. Accumulating $\vJ_{p,i}^T\vJ_{p,i}$ costs $2SM^2$ FLOPs per shard, or $2NM^2/P$ FLOPs over the $N/(PS)$ local shards. In the dual form, the inner loop re-evaluates all $N/S$ global shards for each of the $N/(PS)$ local shards. This amounts to $N^2/(PS)$ single-sample VJPs and outweighs the $3N/P$ VJPs spent on $\vv{d}$, $\tilde{\vJ}_{p,i}$, and $\vJ^T\vv{v}$. The $N^2/(PS^2)$ block products $\tilde{\vJ}_{p,i}\vJ_j^T$ cost $2S^2M$ FLOPs each and $2N^2M/P$ FLOPs in total. Solving the $M \times M$ or $N \times N$ system by LU factorization costs $\mathcal{O}(M^3)$ or $\mathcal{O}(N^3)$ FLOPs. This term does not shrink with $P$, since every device solves the same system. The remaining matrix-vector and element-wise operations, such as $\vJ_{p,i}^T\vv{r}_{p,i}$ and $\vJ_{p,i} \odot \vv{d}^T$, cost $\mathcal{O}(NM/P)$ FLOPs and are bounded by the shard products. The per-device computational complexities are therefore
\begin{equation}\label{eq:computation-complexity}
    \begin{aligned}
        \text{Standard:} & \quad \mathcal{O}\Big(\frac{NF}{P} + \frac{NM^2}{P} + M^3\Big)    \\
        \text{Dual:}     & \quad \mathcal{O}\Big(\frac{N^2F}{PS} + \frac{N^2M}{P} + N^3\Big)
    \end{aligned}
\end{equation}
When $M > N$, the block products of the dual form cost a factor of $M/N$ less than the accumulation of $\vJ^T\vJ$ in the standard form. The extra cost of the dual form comes from evaluating the Jacobian $N/S$ times, which is what it pays to remove the $\mathcal{O}(M^2)$ memory.

\subsubsection{Communication}
The standard form all-reduces $\vJ^T\vJ$ and $\vJ^T\vv{r}$. The dual form all-gathers $\vJ\vv{D}^{-1}\vJ^T$ and all-reduces the two $M$-dimensional vectors $\vv{d}$ and $\vJ^T\vv{v}$. As with memory, the Gram matrix dominates the communication volume, which is $\mathcal{O}(\min(M, N)^2 + M)$ elements. The scalar loss sent over Gloo is negligible.

Besides the model size $M$ and the batch size $N$, the shard size $S$ and the slice size $L$ decide the balance between memory and speed. A larger $S$ reduces the Jacobian re-evaluation of the dual form but costs $\mathcal{O}(SM)$ memory. A larger $L$ improves the utilization of vectorized VJPs at a cost of $\mathcal{O}(LA)$ memory, with the same FLOP count. $S$ and $L$ should therefore be tuned together to fill the device memory that the Gram matrix leaves free.

\section{Evaluation}\label{sec:result}

LeMA changes how an LMA step is computed, not which step is taken. As long as it computes the same update as LMA, it inherits the convergence behavior of LMA, which prior work has studied for neural networks~\cite{hagan1994training,ali2023levenberg,dimarco2025torch-levenberg}. We therefore evaluate correctness and performance rather than training accuracy. We first verify numerically that the dual form matches an FP64 reference update (Section~\ref{sec:result-exactness}). We then measure how far LeMA extends the model size on one GPU (Section~\ref{sec:result-memory}), compare the two forms (Section~\ref{sec:result-forms}), and evaluate strong scaling (Section~\ref{sec:result-strong-scaling}). Finally, we study the shard and slice sizes (Section~\ref{sec:result-sl}) and split-K (Section~\ref{sec:result-split-k}).

Although LeMA implements both the standard and dual forms of LMA, we treat the standard form as the baseline for most experiments because the number of parameters typically exceeds the batch size in practice.

\begin{figure*}[t]
    \centering
    \includegraphics[width=\textwidth]{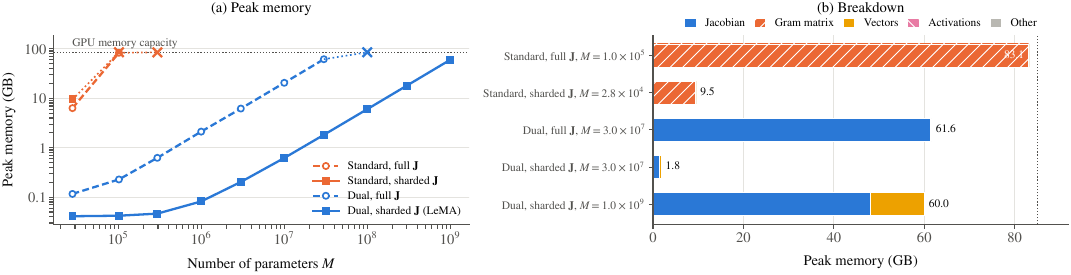}
    \caption{Peak memory of one trial step on one GPU with $N = 256$ and $S = L = 4$. (a) Crosses mark the first model that runs out of memory. (b) Breakdown of each variant at its largest model.}
    \label{fig:memory}
\end{figure*}

\subsection{Experiment Setup}\label{sec:result-setup}

\subsubsection{Hardware and Software}
All experiments are conducted on a server equipped with two Intel Xeon Platinum 8275CL CPUs and eight NVIDIA A100-SXM4 GPUs. Each GPU features 80~GB of memory, a peak memory bandwidth of 2,039~GB/s, and a peak FP32 performance of 19.5~TFLOPS. The GPUs are fully connected via NVLink 3.0, providing 600~GB/s of bidirectional bandwidth per GPU.

We implement LeMA in Python~3.14 using PyTorch~2.14, CUDA~12.6, and NCCL~2.29. All computations are performed in single precision (FP32).

\subsubsection{Datasets and Models}
We use MNIST~\cite{deng2012mnist}, which has 60,000 training and 10,000 test images of $28 \times 28$ pixels. The whole dataset is pinned in host memory~\cite{harris2012optimize}.

To reach the billion-parameter scale on MNIST, we take a convolutional neural network (CNN) with two $3 \times 3$ convolutional layers of 32 and 64 channels, a $2 \times 2$ max pooling, and two fully connected layers, with ReLU after each hidden layer, and widen its hidden fully connected layer to $h$ neurons. The model has $M = 18{,}826 + 9{,}227h$ parameters. We call the models with $h = 10{,}836$ and $108{,}378$ CNN-100M and CNN-1B, with 100,002,598 and 1,000,022,632 parameters. Experiments that need the standard LMA form use smaller models: LeNet-5~\cite{lecun1998gradient} with ReLU and max pooling ($M = 61{,}706$) and a 784-16-10 MLP ($M = 12{,}730$). All models use the CE residual in~\eqref{eq:residuals} with $\epsilon = 10^{-9}$.

\subsubsection{Measurements}
We split the step time into four phases: Jacobian shard evaluation ($\textsc{Jac}$), the Gram block products $\tilde{\vJ}_{p,i}\vJ_j^T$, the NCCL and Gloo collectives including the wait for the slowest process, and the rest, such as $\textsc{Vjp}$, the linear solve, and the loss evaluation of the trial step. To analyze the peak memory footprint, we track all memory allocations during execution. At the exact moment of maximum memory usage, we categorize every live tensor as the Gram matrix, the Jacobian or its shards, $M$-dimensional vectors, activations, or other buffers.

\subsection{Exactness of the Dual Form}\label{sec:result-exactness}

Section~\ref{sec:algorithm-dual-form} proves that the dual form yields the same update as the standard form in exact arithmetic. To verify this equivalence in FP32, we execute a single LeMA step with each form across different models and damping factors.

As a reference independent of LeMA, we explicitly form the full Jacobian $\vJ$ and $\vJ^T\vJ$, clamp $\vv{D}$ with the same rule as both forms (Section~\ref{sec:implementation-standard}), and solve the standard form~\eqref{eq:levenberg-marquardt-residual-equation} directly in FP64 to obtain $\Delta\vtheta_{\text{ref}}$. We evaluate LeNet-5 with batch sizes $N = 1{,}024$ and $8{,}192$ (both smaller than $M$), and an MLP with $N = 4{,}096 < M$ and $N = 16{,}384 > M$. For each setup, we test damping factors $\lambda = 10^{-3}$, $1$, and $10^3$.

Table~\ref{tab:update-equivalence} reports the relative error $\norm{\Delta\vtheta - \Delta\vtheta_{\text{ref}}}_2 / \norm{\Delta\vtheta_{\text{ref}}}_2$. For $\lambda = 10^3$, the errors of both forms are at most $4.4 \times 10^{-6}$. As $\lambda$ decreases, the linear systems become ill-conditioned, and the standard form degrades rapidly. At $\lambda = 10^{-3}$, its error reaches 21\% to 280\%, rendering the update unusable. In contrast, the error of the dual form stays within $1.3 \times 10^{-2}$. To check the reference itself, we also run the dual form of LeMA in FP64. It differs from the reference by at most $5.3 \times 10^{-9}$, and mostly by $10^{-15}$ to $10^{-12}$, so the FP64 reference stays accurate even at $\lambda = 10^{-3}$, and the dual transformation introduces no error beyond rounding. We conclude that the errors in Table~\ref{tab:update-equivalence} arise from the limited FP32 precision, to which the standard form is far more sensitive than the dual form.

\begin{table}[t]
    \caption{Relative Error of the Update Against an FP64 Reference}
    \label{tab:update-equivalence}
    \centering
    \footnotesize
    \setlength{\tabcolsep}{3pt}
    \renewcommand{\arraystretch}{1.2}
    \begin{tabular}{llccc}
    \hline
    Model   & $N$    & $\lambda$ & Standard form          & Dual form              \\
    \hline
    LeNet-5 & 1,024  & $10^{-3}$ & $2.8$                  & $8.8 {\times} 10^{-6}$ \\
            &        & $1$       & $3.7 {\times} 10^{-3}$ & $5.9 {\times} 10^{-6}$ \\
            &        & $10^{3}$  & $4.3 {\times} 10^{-6}$ & $3.6 {\times} 10^{-7}$ \\
    \hline
            & 8,192  & $10^{-3}$ & $2.0$                  & $2.4 {\times} 10^{-4}$ \\
            &        & $1$       & $2.3 {\times} 10^{-3}$ & $1.6 {\times} 10^{-5}$ \\
            &        & $10^{3}$  & $4.4 {\times} 10^{-6}$ & $3.2 {\times} 10^{-7}$ \\
    \hline
    MLP     & 4,096  & $10^{-3}$ & $0.67$                 & $2.3 {\times} 10^{-3}$ \\
            &        & $1$       & $2.8 {\times} 10^{-3}$ & $2.7 {\times} 10^{-5}$ \\
            &        & $10^{3}$  & $4.3 {\times} 10^{-6}$ & $2.6 {\times} 10^{-7}$ \\
    \hline
            & 16,384 & $10^{-3}$ & $0.21$                 & $1.3 {\times} 10^{-2}$ \\
            &        & $1$       & $2.0 {\times} 10^{-3}$ & $1.5 {\times} 10^{-5}$ \\
            &        & $10^{3}$  & $3.4 {\times} 10^{-6}$ & $2.2 {\times} 10^{-7}$ \\
    \hline
\end{tabular}

\end{table}

\subsection{Memory Footprint}\label{sec:result-memory}

We next evaluate the maximum number of parameters each variant supports on a single GPU. We compare four configurations: the standard and dual forms of LMA, each implemented with either a full or a sharded Jacobian. By increasing $h$ from 1 to 108,378, we scale the parameter count $M$ from $2.8 \times 10^4$ to $1.0 \times 10^9$ until each configuration exhausts the GPU memory. All configurations use $N = 256$ and $L = 4$, while the sharded implementations additionally set $S = 4$.

Figure~\ref{fig:memory} shows the peak memory and its breakdown. The standard forms encounter the $M^2$ bottleneck first. For the implementation with the full Jacobian, $\vJ^T\vJ$ and the intermediate data for LU factorization consume all but 0.02~GB of the 83.1~GB footprint at $M \approx 10^5$. As a result, the solver exhausts memory at $M \approx 3 \times 10^5$. The sharded standard form fails even earlier at $M \approx 10^5$ because it retains $\vJ^T\vJ$ for reuse after a rejected step, requiring a third $M \times M$ matrix. The dual form with the full Jacobian avoids the $M^2$ memory limit, but its $N \times M$ Jacobian footprint grows linearly with $M$, reaching 61.6~GB at $M = 3.0 \times 10^7$.

LeMA resolves both of these memory bottlenecks. At $M = 3.0 \times 10^7$, it requires only 1.82~GB, achieving a $34\times$ reduction compared to the dual form with the full Jacobian. For the CNN-1B model, the peak memory of 60.0~GB consists of 48.0~GB for shards and 12.0~GB for vectors. The shards exactly match the $3SM = 12M$ elements predicted in Table~\ref{tab:cost-analysis}, and the vectors amount to three $M$-dimensional vectors, giving $15M$ elements in total. Consequently, LeMA successfully trains a model on a single GPU that is $33\times$ larger than the maximum capacity of the full-Jacobian dual form, and approximately $10^4\times$ larger than that of the standard LMA.

\subsection{Standard Form vs.\ Dual Form}\label{sec:result-forms}

\begin{figure}[t]
    \centering
    \includegraphics[width=\columnwidth]{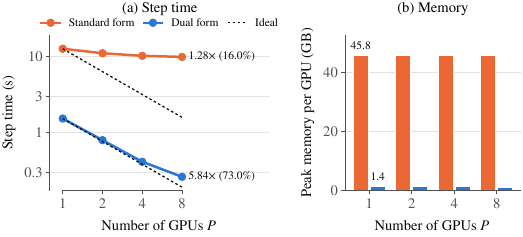}
    \caption{Standard and dual forms for LeNet-5 with $N = 8{,}192 < M = 61{,}706$. Dashed lines show ideal scaling.}
    \label{fig:forms}
\end{figure}

We compare the memory footprint and execution performance of the standard and dual forms. We use LeNet-5 with $N = 8{,}192 < M$. Both forms use $S = 1{,}024$ and $L = 256$ on identical samples.

Figure~\ref{fig:forms} shows that the dual form achieves a smaller memory footprint and faster execution. It outperforms the standard form across all $P$ configurations, delivering an $8.3\times$ speedup on a single GPU and a $37.8\times$ speedup on 8 GPUs. The standard form exhibits poor scaling efficiency. Specifically, on a single GPU it spends 9.23~s of its 12.65~s step time solving the $M \times M$ linear system. Because every GPU solves this system redundantly without parallelization, this 73\% share bounds the speedup to $1.37\times$, and the standard form reaches only $1.28\times$ on 8 GPUs. The dual form scales significantly better. Even on 8 GPUs, it requires only 0.05~s to solve the $N \times N$ system out of a 0.26~s step time, so this unparallelized overhead limits scaling far less, and the dual form reaches $5.84\times$ on 8 GPUs.

The dual form achieves a $33\times$ reduction in memory footprint compared to the standard form. The standard form consumes 45.8~GB per GPU regardless of $P$, dominated by three $M \times M$ matrices: $\vJ^T\vJ$, its damped copy, and the LU factors. In contrast, the dual form requires only 1.37~GB on a single GPU and 1.11~GB on 8 GPUs, which is $33\times$ to $41\times$ less. This footprint decreases as $P$ increases because each GPU computes only $N/P$ rows of the Gram matrix before the all-gather operation.

\subsection{Strong Scaling}\label{sec:result-strong-scaling}

\begin{figure}[t]
    \centering
    \includegraphics[width=\columnwidth]{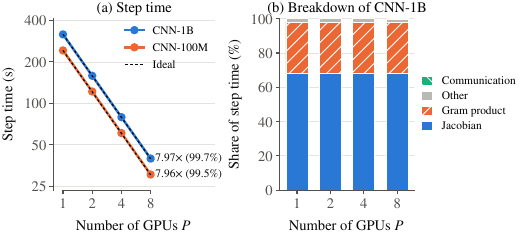}
    \caption{Strong scaling of LeMA with a fixed global batch. (a) Step time, labeled with the speedup and efficiency on 8 GPUs. (b) Phase breakdown of CNN-1B.}
    \label{fig:strong-scaling}
\end{figure}

We scale CNN-100M and CNN-1B from 1 to 8 GPUs with a fixed global batch. Both use the dual form, with $N = 2{,}048$ and $S = L = 32$ for CNN-100M and $N = 256$ and $S = L = 4$ for CNN-1B. These settings are the same for every $P$, so each GPU processes $N/P$ samples.

LeMA scales almost ideally (Figure~\ref{fig:strong-scaling}(a)). On 8 GPUs, the step time of CNN-1B drops from 317.2~s to 39.8~s, a speedup of $7.97\times$ and a parallel efficiency of 99.7\%. CNN-100M reaches $7.96\times$ (99.5\%), and the efficiency stays at or above 99.4\% for every $P$. The peak memory per GPU stays at 60.0~GB for CNN-1B and 39.7~GB for CNN-100M, since~\eqref{eq:memory-complexity} does not depend on $P$.

Figure~\ref{fig:strong-scaling}(b) shows the reason. The Jacobian evaluation and the Gram block products take 68\% and 29\% of the CNN-1B step, and both are split evenly across GPUs. Communication takes only 0.06~s on 8 GPUs, 0.16\% of the step, although it all-reduces two 4~GB vectors. The dual form re-evaluates each Jacobian row $N/S$ times but communicates its $N^2 + 2M$ elements only once per step (Section~\ref{sec:implementation-analysis}). The redundant linear solve takes less than 0.03\% of the step.

\subsection{Shard and Slice Sizes}\label{sec:result-sl}

\begin{figure}[t]
    \centering
    \includegraphics[width=\columnwidth]{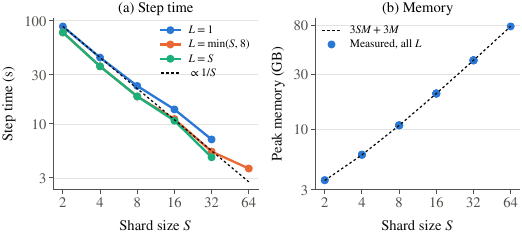}
    \caption{Effect of the shard size $S$ and the slice size $L$ for CNN-100M with $N = 256$ on one GPU.}
    \label{fig:sl-sweep}
\end{figure}

Section~\ref{sec:implementation-analysis} predicts that $S$ trades Jacobian re-evaluation against memory, while $L$ only affects hardware utilization. We train CNN-100M with $N = 256$ on one GPU for $S$ from 2 to 64 and $L$ from 1 to $S$ (Figure~\ref{fig:sl-sweep}).

The step time falls almost in proportion to $1/S$. With $L = S$, it drops from 77.1~s at $S = 2$ to 4.75~s at $S = 32$, $16.2\times$ for a $16\times$ larger shard, and $S = 64$ with $L = 8$ reaches 3.70~s. Larger shards help both dominant phases: the Jacobian is re-evaluated fewer times, and larger blocks raise the arithmetic intensity of the block products at the same FLOP count. The slice size matters much less. At $S = 32$, raising $L$ from 1 to 32 speeds up the step by only $1.5\times$.

The peak memory matches $3SM + 3M$ elements, i.e., the $3SM$ Jacobian shard elements of Table~\ref{tab:cost-analysis} plus three $M$-dimensional vectors, within 0.05~GB and does not depend on $L$, because the activations of a slice are negligible for this model. The only exception is $S = L = 64$, which runs out of memory although it needs the same amount as $S = 64$ with smaller $L$, because about 24~GiB of the reserved memory are too fragmented for its large slice allocations. Since the time scales with $1/S$ and the memory with $S$, $S$ should be as large as the GPU memory allows.

\subsection{Split-K of the Gram Block Products}\label{sec:result-split-k}

\begin{figure}[t]
    \centering
    \includegraphics[width=\columnwidth]{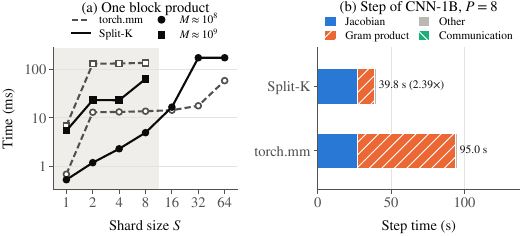}
    \caption{Gram block products with \texttt{torch.mm} and with split-K. (a) One product of two $S \times M$ shards on one GPU. The shaded range is where LeMA applies split-K. (b) Step time of CNN-1B on 8 GPUs.}
    \label{fig:split-k}
\end{figure}

We evaluate split-K (Section~\ref{sec:implementation-dual}) on a single block product and on a full step. The baseline computes each block product with one \texttt{torch.mm} call, which uses cuBLAS. In the microbenchmark, we multiply two random $S \times M$ shards with $M \approx 10^8$ or $10^9$, matching CNN-100M and CNN-1B, and $S$ from 1 to 64, skipping shapes that do not fit in memory. Both variants write into a strided block of a larger matrix, as in Algorithm~\ref{alg:dual-distributed}, and split-K uses its default of 16 chunks per SM (1,728 chunks on an A100). After two warm-up runs, we repeat each product for at least 0.5~s and report the mean time from CUDA events. A block product reads $8SM$ bytes in FP32, so a bandwidth-bound kernel should take time proportional to $S$.

A plain GEMM does not (Figure~\ref{fig:split-k}(a)). From $S = 2$ to $8$, the time of \texttt{torch.mm} stays at 12.9--13.6~ms for $M \approx 10^8$ and 130--135~ms for $M \approx 10^9$ while the data grow $4\times$, so the GEMM is limited by its parallelism. Split-K follows the data volume and is $1.27$--$11.0\times$ faster for $S \le 8$. For the CNN-1B shards with $S = 4$, it cuts the time from 131.3~ms to 22.9~ms ($5.7\times$), which corresponds to 1.40~TB/s, or 69\% of the peak memory bandwidth. For $S \ge 16$, the GEMM already has enough parallelism and split-K is up to $9.8\times$ slower, so LeMA applies split-K only for $S \le 8$. The result is not sensitive to the chunk count, which changes the time by at most 12\% beyond 8 chunks per SM. Split-K changes the products by a relative difference of at most $4.5 \times 10^{-5}$ because of the different FP32 summation order.

For the full step, we train CNN-1B on 8 GPUs with the configuration of Section~\ref{sec:result-strong-scaling}. Each GPU computes $N^2/(PS^2) = 512$ block products per step. Without split-K, they take 66.9~s, 70\% of the 95.0~s step, which matches 512 products of 131~ms each in the microbenchmark. Split-K reduces them by $5.7\times$ to 11.8~s and speeds up the step by $2.39\times$ to 39.8~s (Figure~\ref{fig:split-k}(b)). The peak memory stays at 60.0~GB per GPU.

\section{Conclusion and Future Work}

We presented LeMA, a multi-GPU neural network training framework based on the Levenberg-Marquardt (LMA) algorithm. By utilizing an exact dual formulation that shifts the linear system from the parameter dimension to the batch dimension, alongside efficient partitioning strategies and Split-K optimizations, we successfully scale LMA to train models with billions of parameters. Consequently, LeMA achieves significant speedups over standard LMA implementations.

We identify several limitations in the current design of LeMA, which direct our future work. First, our assumption that each input sample is processed independently prohibits the use of intra-batch operations like Batch Normalization. We plan to extend LeMA to support these layers without introducing significant overhead. Second, the dual form repeatedly evaluates all global Jacobian shards for each local block of the Gram matrix $\vJ \vv{D}^{-1} \vJ^T$. This re-evaluation makes Jacobian computation a primary bottleneck. Since inter-process communication time is currently negligible, future implementations could exchange evaluated shards across GPUs via fast NVLink to avoid this redundant computation. Third, LeMA gathers the full coefficient matrix on every GPU and solves the linear system redundantly. Integrating a distributed solver, such as NVIDIA cuSOLVERMp \cite{cusolvermp}, would allow us to partition the coefficient matrix and solve the system in parallel, yielding further memory and speed improvements. Finally, this work focuses strictly on systems engineering to optimize memory and execution time. We use a basic damping strategy and acceptance criterion, and leave refining them and evaluating convergence on practical deep learning tasks to future work.

\section*{Acknowledgment}
This work was supported by the NVIDIA Academic Grant Program.
The authors used generative AI tools, including Claude and Gemini, to check grammar, improve clarity, and place figures near the text that discusses them. The authors reviewed all changes and take full responsibility for the content.

\bibliographystyle{IEEEtran}
\bibliography{IEEEabrv, references}

\end{document}